\documentclass[10pt,twocolumn,letterpaper]{article}

\usepackage{cvpr}
\usepackage{times}
\usepackage{epsfig}
\usepackage{graphicx}
\usepackage{amsmath}
\usepackage{amssymb}
\usepackage{amsfonts}       

\usepackage{url}
\usepackage[table]{xcolor}
\usepackage{bbm}
\usepackage{booktabs}
\usepackage[T1]{fontenc}
\usepackage{fix-cm}
\usepackage{array}
\usepackage{epsfig}
\usepackage{multirow}

\usepackage{times}
\usepackage{helvet}
\usepackage{courier}
\usepackage{graphicx}
\usepackage{bm}
\usepackage{color}
\usepackage{epstopdf}
\usepackage{caption}
\usepackage{subcaption}
\usepackage{enumitem}
\usepackage{calc}
\usepackage{multirow}
\usepackage{xspace}
\usepackage{booktabs}
\usepackage{mathrsfs}
\usepackage{array}
\usepackage{gensymb}

\usepackage[pagebackref=true,breaklinks=true,letterpaper=true,colorlinks,bookmarks=false]{hyperref}
\usepackage{caption}
\cvprfinalcopy 
\newcommand{\figref}[1]{Fig\onedot~\ref{#1}}
\newcommand{\equref}[1]{Eq\onedot~\eqref{#1}}
\newcommand{\secref}[1]{Sec\onedot~\ref{#1}}
\newcommand{\tabref}[1]{Tab\onedot~\ref{#1}}

\newcommand{\ve}[1]{{\mathbf #1}} 
\newcommand*\rot[1]{\rotatebox{45}{#1}}
\newcommand*\rotf[1]{\rotatebox{35}{#1}}
\newcommand{\hua}[1]{{\mathcal #1}}

\newcommand{\by}[2]{\ensuremath{#1 \! \times \! #2}}

\DeclareRobustCommand\onedot{\futurelet\@let@token\@onedot}
\def\onedot{\ifx\@let@token.\else.\null\fi\xspace}
\def\eg{\emph{e.g.}}

\def\ie{\emph{i.e.}}

\def\etc{\emph{etc}\onedot}

\def\wrt{w.r.t\onedot}

\def\etal{\emph{et al.}}

\def\cvprPaperID{936} 

\ifcvprfinal

\begin{document}

\title{DeLS-3D: Deep Localization and Segmentation with a 3D Semantic Map}

\author{Peng Wang, Ruigang Yang, Binbin Cao, Wei Xu, Yuanqing Lin\\
Baidu Research \\
National Engineering Laboratory for Deep Learning Technology and Applications\\
{\tt\small \{wangpeng54, yangruigang, caobinbin, wei.xu, linyuanqing\}@baidu.com}}

\maketitle

\begin{abstract}
For applications such as augmented reality, autonomous driving, self-localization/camera pose estimation and scene parsing are crucial technologies. In this paper, we propose a unified framework to tackle these two problems simultaneously. The uniqueness of our design is a sensor fusion scheme which integrates camera videos, motion sensors (GPS/IMU), and a 3D semantic map in order to achieve robustness and efficiency of the system. 
Specifically, we first have an initial coarse camera pose obtained from consumer-grade GPS/IMU, based on which a label map can be rendered from the 3D semantic map. Then, the rendered label map and the RGB image are jointly fed into a pose CNN, yielding a corrected camera pose.
In addition, to incorporate temporal information, a multi-layer recurrent neural network (RNN) is further deployed improve the pose accuracy.
Finally, based on the pose from RNN, we render a new label map, which is fed together with the RGB image into a segment CNN which produces per-pixel semantic label.
In order to validate our approach, we build a dataset with registered 3D point clouds and video camera images. Both the point clouds and the images are semantically-labeled. Each video frame has ground truth pose from highly accurate motion sensors.
We show that practically, pose estimation solely relying on images like PoseNet~\cite{Kendall_2015_ICCV} may fail due to street view confusion, and it is important to fuse multiple sensors. Finally, various ablation studies are performed, which demonstrate the effectiveness of the proposed system. In particular, we show that scene parsing and pose estimation are mutually beneficial to achieve a more robust and accurate system.
\end{abstract}
\vspace{-0.5\baselineskip}


\vspace{-1.3\baselineskip}
\section{Introduction}
\vspace{-0.5\baselineskip}
\label{sec:introduction}
In applications like robotic navigation~\cite{ohno2003outdoor} or augment reality~\cite{DBLP:journals/corr/abs-1708-05006}, visual-based 6-DOF camera pose estimation~\cite{campbell2017globally,moreno2008pose,Kendall_2015_ICCV,coskun2017long}, and concurrently parsing each frame of a video into semantically meaningful regions~\cite{ZhaoSQWJ16,WuSH16e,ChenPSA17} efficiently are the key components, which are attracting much attention in computer vision.

\begin{figure*}[t]
\center
\vspace{-1\baselineskip}
\includegraphics[width=0.9\textwidth]{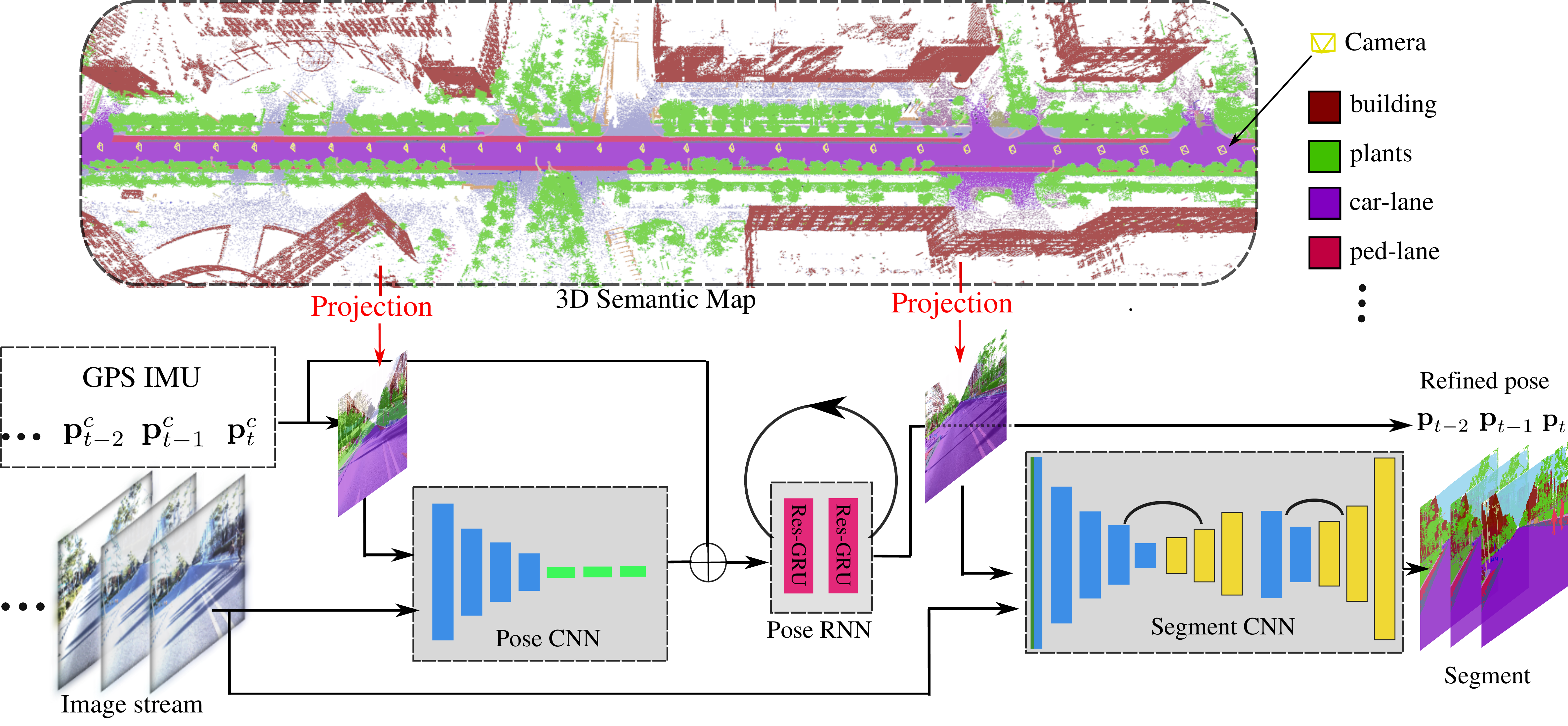}
\caption{System overview. The black arrows show the testing process, and red arrows indicate the rendering (projection) operation in training and inference. The yellow frustum shows the location of cameras inside the 3D map. The input of our system contains a sequence of images and corresponding GPS/IMU signals. The outputs are the semantically segmented images, each with its refined camera pose.}
\label{fig:framework}
\vspace{-1\baselineskip}
\end{figure*}

Currently, most state-of-the-art (SOTA) algorithms try to solve both tasks using solely visual signals.
For camera localization, geometric based methods are relying on visual feature matching, \eg~systems of Perspective-n-Points (PnP)~\cite{haralick1994review,kneip2014upnp,campbell2017globally} when a 3D map and an image is provided, or systems of SLAM~\cite{engel2014lsd,mur2015orb,NewcombeLD11} when there is a video. Such systems are dependent on local appearance, which could fail when confronted with low-texture environments.
Most recently, deep learning based methods, \eg~for either images~\cite{Kendall_2015_ICCV} or videos~\cite{DBLP:journals/corr/ClarkWMTW17}, have been developed, which not only consider hierarchical features, while yielding real-time performance. 
Nevertheless, those methods are good for environments with rich distinguishable features, such as these in the Cambridge landmarks dataset~\cite{Kendall_2015_ICCV}. They could fail for common street views with very similar appearances or even repetitive structures.

For scene parsing, approaches~\cite{ZhaoSQWJ16,ChenPSA17} based on deep fully convolutional network (FCN) with ResNet~\cite{HeZRS15} are the best-performing algorithms for single image input. When the input is video, researchers~\cite{kundu2016feature,zhu2016deep} incorporate the optical flow between consecutive frames, which not only accelerates the parsing, but also improve temporal consistency. Furthermore, for static background, one may use structure-from-motion (SFM) techniques~\cite{wu2011visualsfm} to jointly parse and reconstruct~\cite{kundu2014joint}. However, these methods could be either time-consuming or hard to generalize for applications asking online real-time performance.

In this paper, we aim to solve this camera localization and scene parsing problem jointly from a more practical standpoint. In our system, we assume to have (a) GPS/IMU signal to provide a coarse camera pose estimation; (b) a semantic 3D map for the static environment. The GPS/IMU signals serve as a crucial prior for our deep-learning based pose estimation system. The semantic 3D map, which can synthesize a semantic view for a given camera pose, not only provides strong guidance for scene parsing, but also helps maintain temporal consistency.
%
Our setting is considered on par with the widely used mobile navigation systems, whereas the 2D labeled map is replaced with a 3D semantic map, and we are able to virtually place the navigation signals into the images with more accurate self-localization and scene parsing on-the-fly. 
Promisingly, with the accelerated development of autonomous driving, city-scale 3D semantic maps are being collected and built (such as the TorontoCity dataset~\cite{wang2016torontocity}). Here, we constructed our own data with high quality 3D semantic map, which is captured via a high-accuracy mobile LIDAR device from $Riegl$\footnote{http://www.rieglusa.com/index.html}.

Last but not least, within our deep learning framework, the camera poses and scene semantics are mutually beneficial. The camera poses help establish the correspondences between the 3D semantic map and 2D semantic label map. Conversely, scene semantics could help refine camera poses. Our unified framework yields better results, in terms of both accuracy and speed, for both tasks than doing them individually. In our experiments, using a single Titan Z GPU, the networks in our system estimates the pose in 10ms with accuracy under 1 degree, and segments the image $512 \times 608$ in within 90ms with pixel accuracy around 96$\%$ without model compression, which demonstrates its efficiency and effectiveness.

In summary, the contributions of this paper are:
\begin{itemize}
\vspace{-0.5\baselineskip}
    \setlength{\itemsep}{-2pt}
    \item We propose a deep learning based system for fusing multiple sensors, \ie~RGB images, customer-grad GPS/IMU, and 3D semantic maps, which improves the robustness and accuracy for camera localization and scene parsing.
    \item Camera poses and scene semantics are designed to handle jointly in a unified framework.
    \item We create a dataset from real scenes to fully evaluate our approach. It includes dense 3D semantically labelled point clouds, ground truth camera poses and pixel-level semantic labels of video camera images, which will be released in order to benefit related researches.
\vspace{-0.4\baselineskip}
\end{itemize}

The structure of this paper is organized as follows. We first give an overview of our system in \secref{sub:framework} and talk about related works in \secref{sec:related_work}. In \secref{sec:data_collection}, we describe the uniqueness of our data from the existing outdoor datasets, and introduce our collection and labelling process. 
Then, \secref{sec:localize_and_parsing} presents details of our system. We perform full evaluation quantitatively for both pose estimation and scene parsing in \secref{sec:experiments}, and \secref{sec:conclusion} concludes the paper and points out future directions. 

\vspace{-0.7\baselineskip}
\subsection{Framework}
\vspace{-0.6\baselineskip}
\label{sub:framework}
The framework of our system is illustrated in \figref{fig:framework}. At upper part, a pre-built 3D semantic map is available. During testing, an online stream of images and corresponding coarse camera poses from GPS/IMU are fed into the system. Firstly, for each frame, a semantic label map is rendered out given the input coarse camera pose, which is fed into a pose CNN jointly with the respective RGB image.  The network calculates the relative rotation and translation, and yields a corrected camera pose. To incorporate the temporal correlations, the corrected poses from pose CNN are fed into a pose RNN to further improves the estimation accuracy in the stream.
Last, given the rectified camera pose, a new label map is rendered out, which is fed together with the image to a segment CNN. The rendered label map helps to segment a spatially more accurate and temporally more consistent result for the image stream of video.
In this system, since our data contains ground truth for both camera poses and segments, it can be trained with strong supervision at each end of outputs.

\vspace{-0.6\baselineskip}
\section{Related Work}
\vspace{-0.5\baselineskip}
\label{sec:related_work}
Estimating camera pose and semantic parsing given a video or a single image have long been center problems for computer vision.
Here we summarize the related works in several aspects without enumerating them all due to space limitation.
Notice that our application is autonomous driving and navigation, we therefore focus on outdoor cases with street-view input. 
Localization and general parsing in the wild is beyond the scope of this paper.

\textbf{Camera pose estimation.} Traditionally, localizing an image given a set of 3D points is formulated as a Perspective-$n$-Point (P$n$P) problem~\cite{haralick1994review,kneip2014upnp} by matching feature points in 2D and features in 3D through cardinality maximization. Usually in a large environment, a pose prior is required in order to obtain good estimation~\cite{david2004softposit,moreno2008pose}. Campbell \etal~\cite{campbell2017globally} propose a global-optimal solver which leverage the prior. In the case that geo-tagged images are available, Sattler \etal~\cite{sattler2017large} propose to use image-retrieval to avoid matching large-scale point cloud.
When given a video, temporal information could be further modeled with methods like SLAM~\cite{engel2014lsd} etc, which increases the localization accuracy and speed.

Although these methods are effective in cases with distinguished feature points, they are still not practical for city-scale environment with billions of points, and they may also fail in areas with low texture, repeated structures, and occlusions.
Thus, recently, deep learned features with hierarchical representations are proposed for localization. PoseNet~\cite{Kendall_2015_ICCV,kendall2017geometric} takes a low-resolution image as input, which can estimate pose in 10ms \wrt a feature rich environment composed of distinguished landmarks. LSTM-PoseNet~\cite{hazirbasimage} further captures a global spatial context after CNN features.
Given an video, later works incorporate Bi-Directional LSTM~\cite{DBLP:journals/corr/ClarkWMTW17} or Kalman filter LSTM~\cite{coskun2017long} to obtain better results with temporal information. However, in street-view scenario, considering a road with trees aside, in most cases, no significant landmark appears, which could fail the visual models. Thus, signals from GPS/IMU are a must-have for robust localization in these cases~\cite{vishal2015accurate}, whereas the problem switched to estimating the relative pose between the camera view from a noisy pose and the real pose. For finding relative camera pose of two views, recently, researchers ~\cite{laskar2017camera,ummenhofer2016demon} propose to stack the two images as a network input. In our case, we concatenate the real image with an online rendered label map from the noisy pose, which provides superior results in our experiments.

\textbf{Scene parsing.} For parsing a single image of street views (e.g., these from CityScapes~\cite{Cordts2016Cityscapes}), most state-of-the-arts (SOTA) algorithms are designed based on a FCN~\cite{WuSH16e} and a multi-scale context module with dilated convolution~\cite{ChenPSA17}, pooling~\cite{ZhaoSQWJ16}, CRF~\cite{higherordercrf_ECCV2016}, or spatial RNN~\cite{byeon2015scene}. However, they are dependent on a ResNet~\cite{HeZRS15} with hundreds of layers, which is too computationally expensive for applications that require real-time performance. 
Some researchers apply small models~\cite{PaszkeCKC16} or model compression~\cite{ZhaoQSSJ17} for acceleration, with the cost of reduced accuracy.
When the input is a video, spatial-temporal informations are jointly considered, Kundu \etal~\cite{kundu2016feature} use 3D dense CRF to get temporally consistent results. Recently, optical flow~\cite{dosovitskiy2015flownet} between consecutive frames is computed to transfer label or features~\cite{gadde2017semantic,zhu2016deep} from the previous frame to current one.  In our case, we connect consecutive video frames through 3D information and camera poses, which is a more compact representation for static background. 
In our case, we propose the projection from 3D maps as an additional input, which alleviates the difficulty of scene parsing solely from image cues. Additionally, we adopt a light weighted network from DeMoN~\cite{ummenhofer2016demon} for inference efficiency.

\textbf{Joint 2D-3D for video parsing.} Our work is also related to joint reconstruction, pose estimation and parsing~\cite{kundu2014joint,hane2013joint} through embedding 2D-3D consistency.
 Traditionally, reliant on structure-from-motion (SFM)~\cite{hane2013joint} from feature or photometric matching, those methods first reconstruct a 3D map, and then perform semantic parsing over 2D and 3D jointly, yielding geometrically consistent segmentation between multiple frames.
 Most recently, CNN-SLAM~\cite{tateno2017cnn} replaces traditional 3D reconstruction module with a single image depth network, and adopts a segment network for image parsing.
 However, all these approaches are processed off-line and only for static background, which do not satisfy our online setting. Moreover, the quality of a reconstructed 3D model is not comparable with the one collected with a 3D scanner. 

\vspace{-0.5\baselineskip}
\section{Dataset}
\vspace{-0.3\baselineskip}

\label{sec:data_collection}
\begin{table}[b]
\center
\vspace{-1\baselineskip}
\fontsize{7}{7}\selectfont
\hspace*{-0.1cm}
\setlength\tabcolsep{1.5pt}
\begin{tabular}{lcccc}
\toprule[0.13em]
Dataset & Real data & Camera pose & 3D semantic map & Video per-pixel label   \\
\hline
\multicolumn{1}{l|}{CamVid~\cite{brostow2009semantic}}     &\checkmark        & -              & -              &  -   \\
\multicolumn{1}{l|}{KITTI~\cite{geiger2012we}}      &\checkmark  & \checkmark     & sparse points  & -   \\
\multicolumn{1}{l|}{CityScapes~\cite{Cordts2016Cityscapes}} &\checkmark  & -              &  -             & selected frames  \\
\multicolumn{1}{l|}{Toronto~\cite{wang2016torontocity}}    &\checkmark  & \checkmark     & 3d building $\&$ road & selected pixels \\
\hline
\multicolumn{1}{l|}{Synthia~\cite{RosCVPR16}}    & -          & \checkmark     & -       &\checkmark     \\
\multicolumn{1}{l|}{P.F.B.~\cite{richter2017playing}} &-   & \checkmark     & -     &\checkmark  \\
\hline
\multicolumn{1}{l|}{Ours}              & \checkmark &\checkmark    &dense point cloud  & \checkmark    \\
\toprule[0.13 em]
\vspace{-1.1\baselineskip}
\end{tabular}
\caption{Compare our dataset with the other related outdoor street-view datasets for our task. `Real data' means whether the data is collected from the physical world.
`3D semantic map' means whether it contains a 3D map of scenes with semantic label. `Video per-pixel label' means whether it has per-pixel semantic label.}
\label{tbl:data}
\vspace{-1.0\baselineskip}
\end{table}

\textbf{Motivation.}
As described in the \secref{sub:framework}, our system is designed to work with available 3D motion sensors and a semantic 3D map.
However, public outdoor datasets such as KITTI and CityScapes do not contain such information, in particular the 3D map. The TortoroCity dataset~\cite{wang2016torontocity} may be used while is not open to public yet. As summarized in \tabref{tbl:data}, we list several key requirements to perform our experiments, which none of current existing datasets fully satisfy.  

\begin{figure}[t]
\begin{center}
\includegraphics[width=\linewidth]{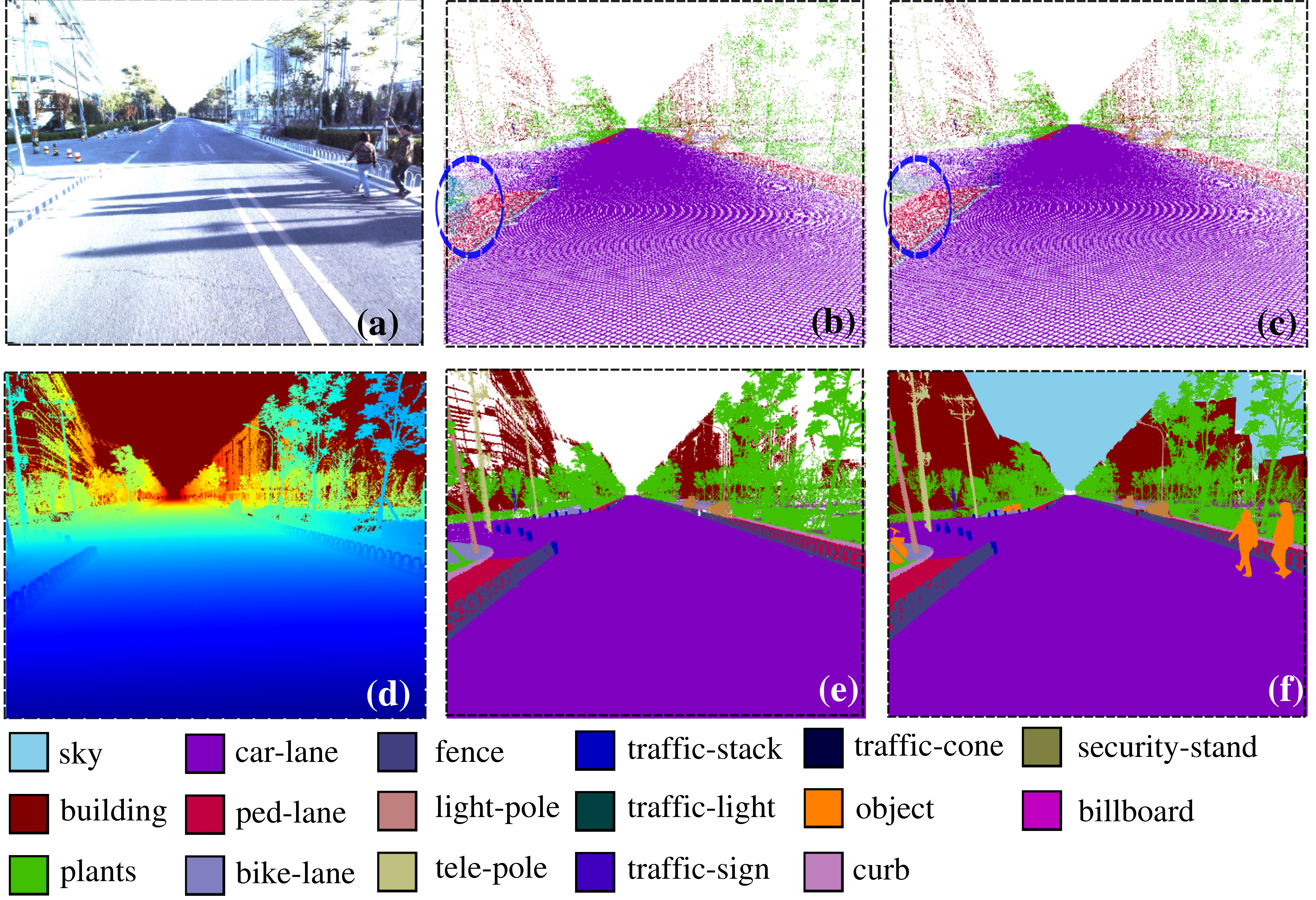}
\vspace{-1.7\baselineskip}
\end{center}
	\caption{An example of our collected street-view dataset. (a) Image. (b) Rendered label map with 3D point cloud projection, with an inaccurate moving object (rider) circled in blue. (c) Rendered label map with 3D point cloud projection after points with low temporal consistency being removed. (d) $\&$ (e) Rendered depth map of background and rendered label map after class dependent splatting in 3D point clouds (\secref{sub:render}). (f) Merged label map with missing region in-painted, moving objects and sky. An example of our labeled videos is shown by an online \href{https://youtu.be/M6lhkzKFEhA}{video}.}
\label{fig:data}
\vspace{-1.3\baselineskip}
\end{figure}

\textbf{Data collection.}
We use a mobile LIDAR scanner from $Riegl$ to collect point clouds of the static 3D map with high granularity. As shown in \figref{fig:data}(a). The captured point cloud density is much higher than the Velodyne\footnote{http://www.velodynelidar.com/} used by KITTI~\cite{geiger2012we}.
Different from the sparse Velodyne LIDAR, our mobile scanner utilizes two laser beams to scan vertical circles. As the acquisition vehicle moves, it scans its surroundings as a push-broom camera. However, moving objects, such as vehicles and pedestrians, could be compressed, expanded, or completely missing in the captured point clouds.
In order to eliminate these inaccurate moving objects (circled in blue at \figref{fig:data}(b)), we conduct three steps:
1) scan the same road segment multiple rounds; 2) align and fuse those point clouds; 3) remove the points with low temporal consistency.
Formally, the condition to kept a point $\ve{x}$ in round $j$ is,
{\vspace{-0.5\baselineskip}
\begin{align}
\sum_{i=0}^{r}{\mathbbm{1}(\exists~\ve{x}_i \in \hua{P}_i~s.t.~\|\ve{x}_i - \ve{x}_j\| < \epsilon_d )} / r \geq \delta
\end{align}
}
where $\delta = 0.6$ and $\epsilon_d = 0.025m$ in our setting, and $\mathbbm{1}()$ is an indicator function. 
We keep the remained point clouds as a static background $\hua{M}$ for further labelling.

For video capturing, we use two frontal cameras with a resolution of \by{2048}{2432}. The whole system including the LIDAR scanner and cameras is well calibrated.

\textbf{2D and 3D labeling}
In order to have semantic labelling of each video frame, we handle static background, static objects (\eg parked vehicles that could be well recognized in point clouds), and moving objects separately.
Firstly, for static background, we directly perform labelling on the 3D point clouds $\hua{M}$ which are then projected to images, yielding labelled background for all the frames.
Specifically, we over-segment the point clouds into point clusters based on spatial distances and normal directions, and then label each cluster of points manually.
Second, for static objects in each round, we prune out the points of static background, and label the remaining points of the objects.
Thirdly, after 3D-2D projection, only moving objects remain unlabeled. Here, we adopt an active labelling strategy, by first training an object segmentation model using a SOTA algorithm~\cite{WuSH16e}, and then refining the masks of moving objects manually.

As shown in \figref{fig:data}(c), the labels obtained from above three steps are still not perfect. There are some unlabeled pixels that could be caused by missing points or reflection. We handle such issues by using the splatting technique in computer graphics, which turns each point into a small square as discussed in \secref{sub:render} (\figref{fig:data}(e)). The results are further refined to generate the final labels (\figref{fig:data}(f)).
With such a strategy, we can greatly increase labelling efficiency and accuracy for video frames. For example, it could be very labor-intensive to label texture-rich regions like trees and poles, especially when occlusion happens. We provide the labelled video in our supplementary materials for readers who are interested. 

Finally, due to space limitation, we elaborate the whole acquisition system, data collection and labelling process with an extended dataset paper~\cite{huang2018cvprw}, called ApolloScape~\footnote{http://apolloscape.auto/}, where a larger dataset with more labelled objects is collected and organized after our submission. 
Nevertheless, in this paper, we experimented with a preliminary version of that dataset, which will be released separately for reproducibility of our results.

\vspace{-0.5\baselineskip}
\section{Localizing camera and scene parsing.}
\vspace{-0.3\baselineskip}
\label{sec:localize_and_parsing}
As shown in \secref{sub:framework}, our full system is based on a semantic 3D map and deep networks. In the following, we will first describe how a semantic label map is rendered from the 3D, then discuss the details of our network architectures and the loss functions to train the whole system.

\subsection{Render a label map from a camera pose.}
\label{sub:render}
Formally, given a 6-DOF camera pose $\ve{p} = [\ve{q}, \ve{t}] \in SE(3)$, where $\ve{q} \in SO(3)$ is the quaternion representation of rotation and $\ve{t} \in \mathbbm{R}^3$ is translation, a label map can be rendered from the semantic 3D map, where z-buffer is applied to find the closest point at each pixel.

In our setting, the 3D map is a point cloud based environment. 
Although the density of the point cloud is very high (one point per 25mm within road regions), when the 3D points are far away from the camera, the projected labels could be sparse, \eg regions of buildings shown in \figref{fig:data}(c).
Thus for each point in the environment, we adopt the point splatting technique, by enlarging the 3D point to a square where the square size is determined by its semantic class. Formally, for a 3D point $\ve{x}$ belonging a class $c$, its square size $s_c$ is set to be proportional to the class' average distance to the camera. Formally,
{\vspace{-0.3\baselineskip}
\begin{align}
\label{eq:square_size}
s_c \propto \frac{1}{|\hua{P}_c|}\sum_{\ve{x}\in \hua{P}_c} \min_{\ve{t}\in\hua{T}} d(\ve{x}, \ve{t})
\end{align}
}
where $\hua{P}_c$ is the set of 3D points belong to class $c$, and $\hua{T}$ is the set of ground truth camera poses. 
Then, given the relative square size between different classes, we define an absolute range to obtain the actual square size for splatting. This is non-trivial since too large size will result in dilated edges, while too small size will yield many holes. In our experiments, we set the range as $[0.025, 0.05]$, and find that it provides the highest visual quality.

As shown in \figref{fig:data}(e), invalid values in-between those projected points are well in-painted, meanwhile the boundaries separating different semantic classes are also well preserved. Later, we insert such a rendered label map for the pose CNN and segment CNN, which guides the network to localize the camera and parse the image.

\begin{figure}[t]
\begin{center}
\includegraphics[width=.8\linewidth]{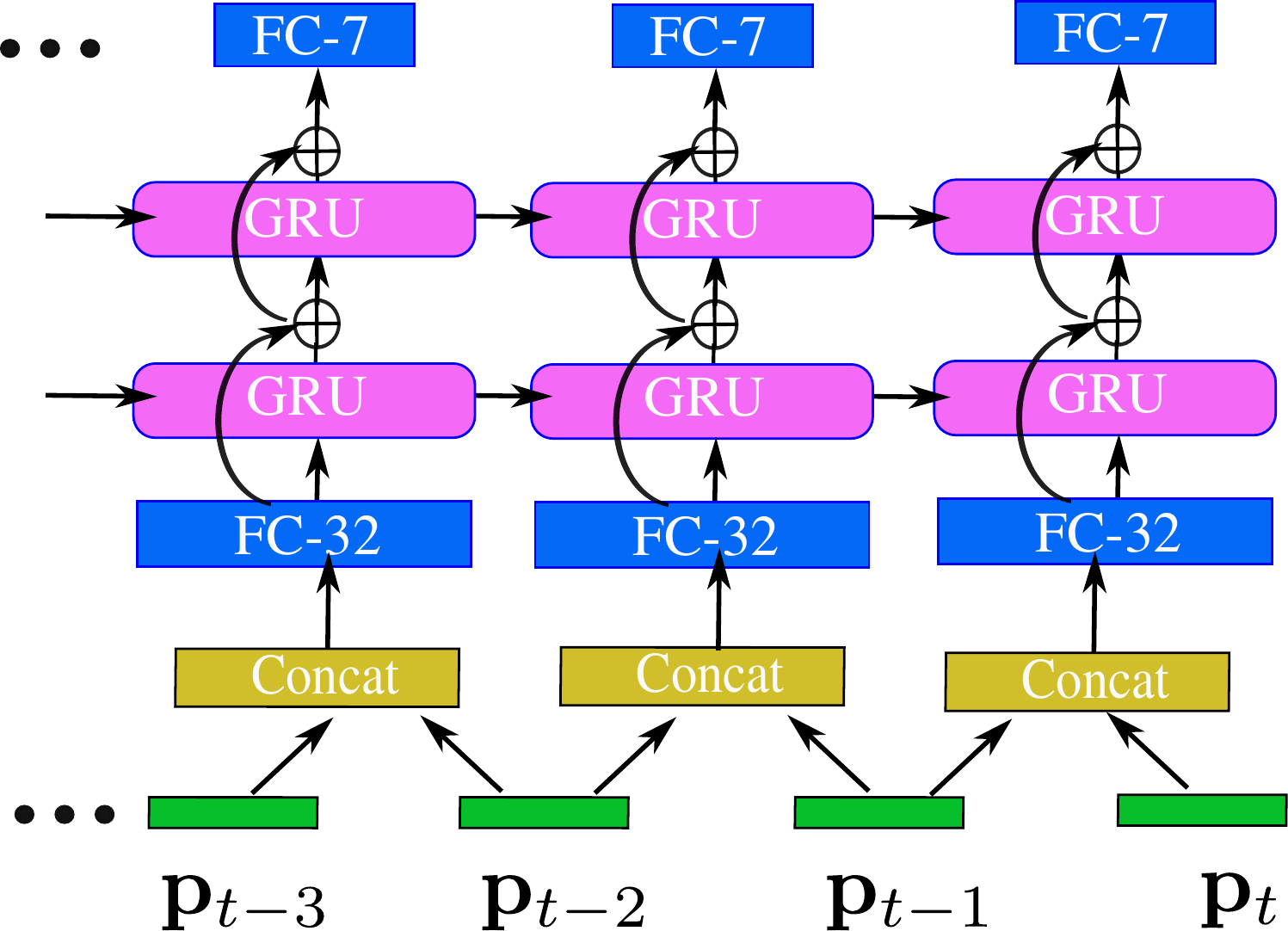}
\end{center}
   \caption{The GRU RNN network architecture for modeling a sequence of camera poses.}
\label{fig:rnn}
\vspace{-1.3\baselineskip}
\end{figure}

\vspace{-0.3\baselineskip}
\subsection{Camera localization with motion prior}
\vspace{-0.2\baselineskip}

\textbf{Translation rectification with road prior.} One common localization priori for navigation is to use the 2D road map, by constraining the GPS signals inside the road regions. We adopt a similar strategy, since once the GPS signal is out of road regions, the rendered label map will be totally different from the street-view of camera, and no correspondence can be found by the network.

To implement this constraint, firstly we render a 2D road map image with a rasterization grid of $0.05m$ from our 3D semantic map by using only road points, \ie points belong to car-lane, pedestrian-lane and bike-lane \etc
Then, at each pixel $[x, y] \in \mathbbm{Z}^2$ in the 2D map, an offset value $\ve{f}(x, y)$ is pre-calculated indicating its 2D offset to the closest pixel belongs to road through the breath-first-search (BFS) algorithm efficiently.

During online testing, given a noisy translation $\ve{t}=[t_x, t_y, t_z]$, we can find the closest road points \wrt $\ve{t}$ using $[t_x, t_y] + \ve{f}(\lfloor t_x \rfloor, \lfloor t_y \rfloor)$ from our pre-calculated offset function. Then, a label map is rendered based on the rectified camera pose, which is fed to pose CNN.

\begin{figure*}[t]
\center
\vspace{-0.6\baselineskip}
\includegraphics[width=0.95\textwidth]{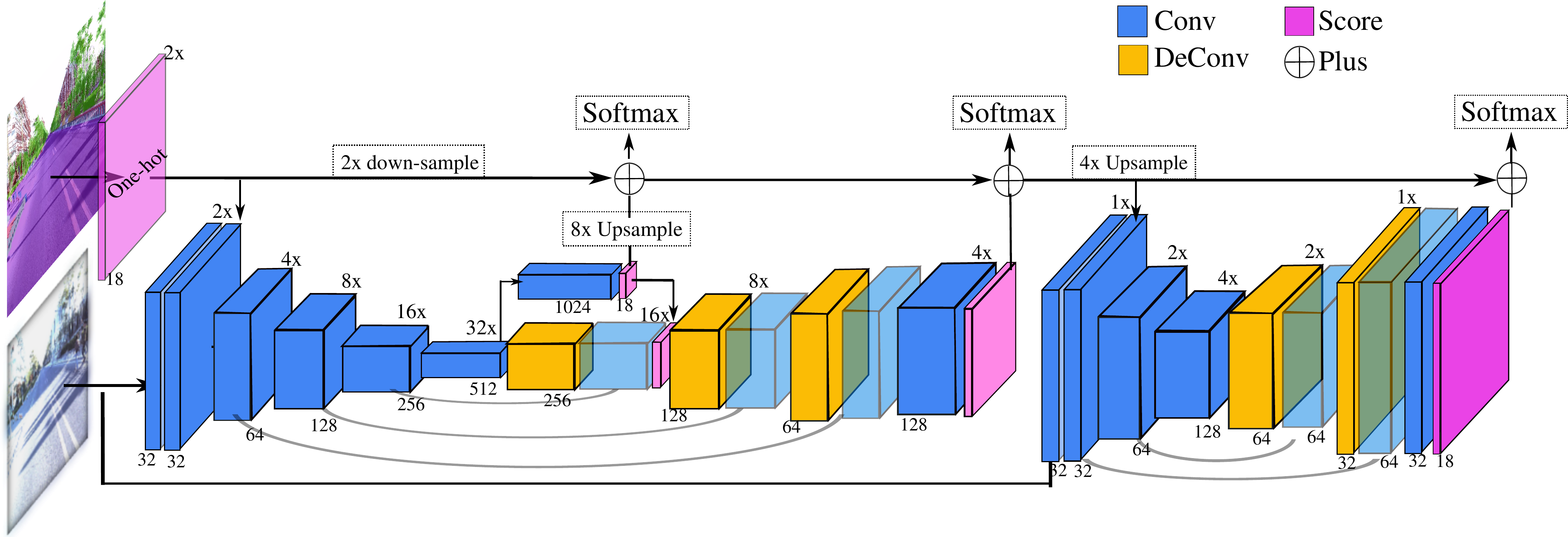}
\caption{Architecture of the segment CNN with rendered label map as a segmentation priori. At bottom of each convolutional block, we show the filter size, and at top we mark the downsample rates of each block \wrt the input image size. The 'softmax' text box indicates the places a loss is calculated. Details are in \secref{subsec:parsing}.}
\label{fig:segnet}
\vspace{-1.35\baselineskip}
\end{figure*}

\textbf{CNN-GRU pose network architecture.}
As shown in \figref{fig:framework}, our pose networks contain a pose CNN and a pose GRU-RNN. Particularly,
the CNN of our pose network takes as inputs an image $\ve{I}$ and the rendered label map $\ve{L}$ from corresponding coarse camera pose $\ve{p}_i^c$. It outputs a 7 dimension vector $\hat{\ve{p}}_i$ representing the relative pose between the image and rendered label map, and we can get a corrected pose \wrt the 3D map by $\ve{p}_i = \ve{p}_i^c + \hat{\ve{p}_i}$.
For the network architecture of pose CNN, we follow the design of DeMoN~\cite{ummenhofer2016demon}, which has large kernel to obtain bigger context while keeping the amount of parameters and runtime manageable. The convolutional kernel of this network consists a pair of 1D filters in $y$ and $x$-direction, and the encoder gradually reduces the spatial resolution with stride of 2 while increasing the number of channels. We list the details of the network in our implementation details at \secref{sec:experiments}.

Additionally, since the input is a stream of images, in order to model the temporal dependency,
after the pose CNN, a multi-layer GRU with residual connection~\cite{wu2016google} is appended.
More specifically, we adopt a two layer GRU with 32 hidden states as illustrated in \figref{fig:rnn}. It includes high order interaction beyond nearby frames, which is preferred for improve the pose estimation performance.
In traditional navigation applications of estimating 2D poses,  Kalman filter~\cite{kalman1960new} is commonly applied by assuming either a constant velocity or acceleration.
In our case, because the vehicle velocity is unknown, transition of camera poses is learned from the training sequences, and in our experiments we show that the motion predicted from RNN is better than using a Kalman filter with a constant speed assumption, yielding further improvement over the estimated ones from our pose CNN.

\textbf{Pose loss.}
Following the PoseNet~\cite{kendall2017geometric}, we use the geometric matching loss for training, which avoids the balancing factor between rotation and translation.
Formally, given a set of point cloud in 3D $~\hua{P}=\{\ve{x}\}$, and the loss for each image is written as,
{\vspace{-0.5\baselineskip}
\begin{align}
L(\ve{p}, \ve{p}^*) = \sum_{\ve{x} \in \hua{P}}\omega_{l_\ve{x}}|\pi(\ve{x}, \ve{p}) - \pi(\ve{x}, \ve{p}^*)|_2
\label{eq:proj_loss}
\end{align}
}
where $\ve{p}$ and $\ve{p}^*$ are the estimated pose and ground truth pose respectively. $\pi()$ is a projective function that maps a 3D point $\ve{x}$ to 2D image coordinates. $l_\ve{x}$ is the semantic label of $\ve{x}$ and $\omega_{l_\ve{x}}$ is a weight factor dependent on the semantics. Here, we set stronger weights for point cloud belong to certain classes like traffic light, and find it helps pose CNN to achieve better performance.
In~\cite{kendall2017geometric}, only the 3D points visible to the current camera are applied to compute this loss to help the stableness of training. However, the amount of visible 3D points is still too large in practical for us to apply the loss.
Thus, we pre-render a depth map for each training image with a resolution of $256 \times 304$ using the ground truth camera pose, and use the back projected 3D points from the depth map for training.

\vspace{-0.3\baselineskip}
\subsection{Video parsing with pose guidance}
\vspace{-0.3\baselineskip}
\label{subsec:parsing}
Having rectified pose at hand, one may direct render the semantic 3D world to the view of a camera, yielding a semantic parsing of the current image. However, the estimated pose is not perfect, fine regions such as light poles can be completely misaligned. Other issues also exist. For instance, many 3D points are missing due to reflection, \eg regions of glasses, and points can be sparse at long distance. Last, dynamic objects in the input cannot be represented by the projected label map, yielding incorrect labelling at corresponding regions. Thus, we propose an additional segment CNN to tackle these issues, while taking the rendered label map as segmentation guidance. 

\textbf{Segment network architecture.} As discussed in \secref{sec:related_work}, heavily parameterized networks such as ResNet are not efficient enough for our online application. Thus, as illustrated in \figref{fig:segnet}, our segment CNN is a light-weight network containing an encoder-decoder network and a refinement network, and both have similar architecture with the corresponding ones used in DeMoN~\cite{ummenhofer2016demon} including 1D filters and mirror connections. However, since we have a segment guidance from the 3D semantic map, we add a residual stream (top part of \figref{fig:segnet}), which encourages the network to learn the differences between the rendered label map and the ground truth. In~\cite{pohlen2016full}, a full resolution stream is used to keep spatial details, while here, we use the rendered label map to keep the semantic spatial layout.

Another notable difference for encoder-decoder network from DeMoN is that for network inputs, shown in \figref{fig:segnet}, rather than directly concatenate the label map with input image, we transform the label map to a score map through one-hot operation, and embed the score of each pixel to a 32 dimensional feature vector. 
Then, we concatenate this feature vector with the first layer output from image, where the input channel imbalance between image and label map is alleviated, which is shown to be useful by previous works~\cite{eigen2015predicting}.
 For refinement network shown in \figref{fig:segnet}, we use the same strategy to handle the two inputs. 
 Finally, the segment network produces a score map, yielding the semantic parsing of the given image.

We train the segment network first with only RGB images, then fine-tune the network by adding the input of rendered label maps. This is because our network is trained from scratch, therefore it needs a large amount of data to learn effective features from images. However, the rendered label map from the estimated pose has on average 70$\%$ pixel accuracy, leaving only 30$\%$ of pixels having effective gradients. This could easily drive the network to over fit to the rendered label map, while slowing down the process towards learning features from images. Finally, for segmentation loss, we use the standard softmax loss, and add intermediate supervision right after the outputs from both the encoder and the decoder as indicated in \figref{fig:segnet}.

\vspace{-1.0\baselineskip}
\section{Experiments}
\vspace{-0.4\baselineskip}
\label{sec:experiments}
We perform all experiments using our collected dataset, and evaluate multiple system settings for pose estimation and segmentation to validate each component.
For GPS and IMU signal, despite we have multiple scans for the same road segments, it is still very limited for training. Thus, follow~\cite{vishal2015accurate}, we simulate noisy GPS and IMU by adding random perturbation $\epsilon$ \wrt the ground truth pose following uniform distributions. 
Specifically, translation and rotation noise are set as $\epsilon_t \sim U(0, 7.5m)$ and $\epsilon_r \sim U(0^{\circ}, 15^{\circ})$ respectively. 
We refer to realistic data~\cite{lee2015gps} for setting the noisy range of simulation.

\textbf{Datasets.} In this paper, our acquisition vehicle scans two sites at Beijing in China yielding two datasets. 
The first one is inside a technology park, named zhongguancun park (Zpark), and we scanned 6 rounds during different daytimes. The 3D map generated has a road length around 3$km$, and the distance between consecutive frames is around 5$m$ to 10$m$. We use 4 rounds of the video camera images for training and 2 for testing, yielding 2242 training images and 756 testing images. 
The second one we scanned 10 rounds and 4km near a lake, named daoxianghu lake (Dlake), and the distance between consecutive frames is around 1$m$ to 3$m$. 
We use 8 rounds of the video camera images for training and 2 for testing, yielding 17062 training images and 1973 testing images. 
The semantic classes of the two datasets are shown in \tabref{tbl:segment}. We will release the two datasets separately. 

\textbf{Implementation details.} To quickly render from the 3D map, we adopt OpenGL to efficiently render a label map with the z-buffer handling. A 512 $\times$ 608 image can be generated in 70ms with a single Titan Z GPU, which is also the input size for both pose CNN and segment CNN. 
For pose CNN, the filter sizes of all layers are $\{32, 32, 64, 128, 256, 1024, 128, 7\}$, and the forward speed for each frame is 9ms. For pose RNN, we sample sequences with length of 100 from our data for training, and the speed for each frame is 0.9ms on average.
For segment CNN, we keep the size the same as input, and the forward time is 90ms. 
Both of the network is learned with 'Nadam' optimizer~\cite{dozat2016incorporating} with a learning rate of $10^{-3}$. We sequentially train these three models due to GPU memory limitation.
Specifically, for pose CNN and segment CNN, we stops at 150 epochs when there is no performance gain, and for pose RNN, we stops at 200 epochs. For data augmentation, we use the imgaug\footnote{https://github.com/aleju/imgaug} library to add lighting, blurring and flipping variations. We keep a subset from training images for validating the trained model from each epoch, and choose the model performing best for evaluation.

For testing, since input GPS/IMU varies every time, \ie~$\ve{p}_t^c=\ve{p}^*+\epsilon$, we need to have a confidence range of prediction for both camera pose and image segment, in order to verify the improvement of each component we have is significant. Specifically, we report the standard variation of the results from a 10 time simulation to obtain the confidence range. Finally, we implement all the networks by adopting the MXNet~\cite{ChenLLLWWXXZZ15} platform.

For pose evaluation, we use the median translation offset and median relative angle~\cite{Kendall_2015_ICCV}. For evaluating segment, we adopt the commonly used pixel accuracy (Pix. Acc.), mean class accuracy (mAcc.) and mean intersect-over-union (mIOU) as that from~\cite{WuSH16e}.
\begin{table}
\vspace{-0\baselineskip}
\center
\fontsize{6.5}{7}\selectfont
\hspace*{-0.23cm}
\begin{tabular}{lllll}
\toprule[0.1 em]
Data & Method & Trans (m) $\downarrow$ & Rot ($\circ$)$\downarrow$  & Pix. Acc($\%$)$\uparrow$ \\
\hline
\parbox[t]{2mm}{\multirow{6}{*}{\rotatebox[origin=c]{90}{Zpark}}} & Noisy pose & 3.45 $\pm$ 0.176 & 7.87 $\pm$ 1.10 & 54.01 $\pm$ 1.5 \\
& Pose CNN w/o semantic & 1.355 $\pm$ 0.052  & 0.982 $\pm$ 0.023 & 70.99 $\pm$ 0.18 \\
& Pose CNN w semantic & 1.331 $\pm$ 0.057  & 0.727 $\pm$ 0.018 & 71.73 $\pm$ 0.18  \\
& Pose RNN w/o CNN & 1.282 $\pm$ 0.061  & 1.731 $\pm$ 0.06 &  68.10 $\pm$ 0.32 \\
& Pose CNN w KF & 1.281 $\pm$ 0.06  & 0.833 $\pm$ 0.03 & 72.00 $\pm$ 0.17  \\
& Pose CNN-RNN  & \textbf{1.005} $\pm$ 0.044  & \textbf{0.719} $\pm$ 0.035  & \textbf{73.01} $\pm$ 0.16  \\
\toprule[0.1 em]
\hline
\parbox[t]{1mm}{\multirow{3}{*}{\rotatebox[origin=c]{90}{Dlake}}} & Pose CNN w semantic & 1.667 $\pm$ 0.05 & 0.702 $\pm$ 0.015 & 87.83 $\pm$ 0.017 \\
& Pose RNN w/o CNN & 1.385 $\pm$ 0.057 & 1.222 $\pm$ 0.054 & 85.10 $\pm$ 0.03\\
& Pose CNN-RNN  & \textbf{0.890} $\pm$ 0.037  & \textbf{0.557}$\pm$ 0.021 & \textbf{88.55} $\pm$ 0.13  \\
\toprule[0.1 em]
\end{tabular}
\caption{Compare the accuracy of different settings for pose estimation from the two datasets.
Noisy pose indicates the noisy input signal from GPS, IMU, and 'KF' means kalman filter.
The number after $\pm$ indicates the standard deviation (S.D.) from 10 simulations. $\downarrow \& \uparrow$ means lower the better and higher the better respectively. 
We can see the improvement is statistically significant.}
\label{tbl:pose}
\vspace{-2.5\baselineskip}
\end{table}

\textbf{Pose Evaluation.}
In \tabref{tbl:pose}, we show the performance of estimated translation $\ve{t}$ and rotation $\ve{r}$ from different model variations. 
We first directly follow the work of PoseNet~\cite{Kendall_2015_ICCV,kendall2017geometric}, and use their published code and geometric loss (\equref{eq:proj_loss}) to train a model on Zpark dataset. 
Due to scene appearance similarity of the street-view, we did not obtain a reasonable model, \ie~results better than the noisy GPS/IMU signal.
At the 1st row, we show the median error of GPS and IMU from our simulation. 
At the 2nd row, by using our pose CNN, the model can learn good relative pose between camera and GPS/IMU, which significantly reduces the error (60$\%$ for $\ve{t}$, 85$\%$ for $\ve{r})$. 
By adding semantic cues, \ie road priori and semantic weights in \equref{eq:proj_loss}, the pose errors are further reduced, especially for rotation (from $0.982$ to $0.727$ at the 3rd row). In fact, we found the most improvement is from semantic weighting, while the road priori helps marginally. In our future work, we would like to experiment larger noise and more data variations, which will better validate different cues.

\begin{figure*}[!htbp]
\center
\vspace{-0.3\baselineskip}
\includegraphics[width=0.99\textwidth]{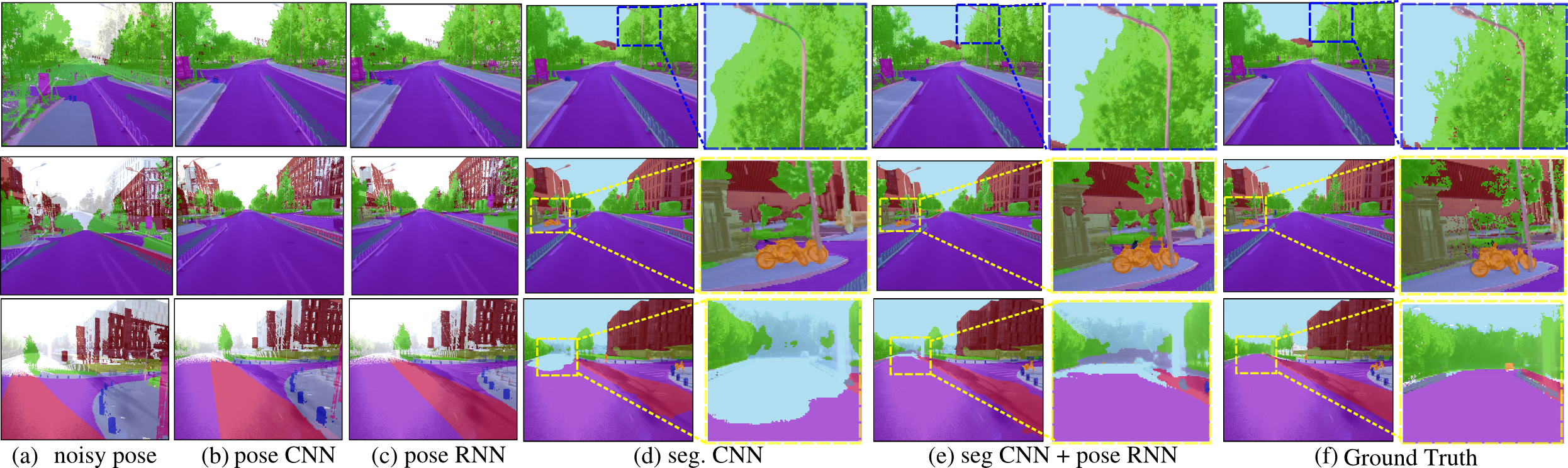}
   \caption{Results from each intermediate stage out of the system over Zpark dataset. Label map is overlaid with the image. Improved regions are boxed and zoomed out (best in color). More results are shown in the online videos for \href{https://youtu.be/HNPQVtgpjbE}{Zpark} and \href{https://youtu.be/ApyqPnvmJAs}{Dlake}.}
\label{fig:results}
\vspace{-1.2\baselineskip}
\end{figure*}

For evaluating an video input, we setup a baseline of performing RNN directly on the GPS/IMU signal, and as shown at 'Pose RNN w/o CNN', the estimated $\ve{t}$ is even better than pose CNN, while $\ve{r}$ is comparably much worse. This meets our expectation since the speed of camera is easier to capture temporally than rotation. Another baseline we adopt is performing Kalman filter~\cite{kalman1960new} to the output from Pose CNN by assuming a constant speed which we set as the averaged speed from training sequences. As shown at 'Pose CNN w KF', it does improve slightly for translation, but harms rotation, which means the filter over smoothed the sequence. Finally when combining pose CNN and RNN, it achieves the best pose estimation both for $\ve{t}$ and $\ve{r}$. We visualize some results at \figref{fig:results}(a-c).
Finally at bottom of \tabref{tbl:pose}, we list corresponding results on Dlake dataset, which draws similar conclusion with that from Zpark dataset.


\begin{table*}[t]
\center
\vspace{-0.5\baselineskip}
\fontsize{6.5}{7}\selectfont
\setlength\tabcolsep{1.5pt}
\begin{tabular}{llcccccccccccccccccccc}
\vspace{-1.0\baselineskip}
Data & \multicolumn{1}{c}{Method} & \rot{mIOU}  & \rot{Pix. Acc}   & \rot{sky} & \rot{car-lane} & \rot{ped-lane} & \rot{bike-lane} & \rot{curb} & \rot{$t$-cone} & \rot{$t$-stack} & \rot{$t$-fence} & \rot{light-pole} & \rot{$t$-light} & \rot{tele-pole} & \rot{$t$-sign} & \rot{billboard} & \rot{temp-build} & \rot{building} & \rot{sec.-stand} & \rot{plants} & \rot{object} \\
\hline
\parbox[t]{2mm}{\multirow{6}{*}{\rotatebox[origin=c]{90}{Zpark}}} & 
ResNet38~\cite{WuSH16e}     &64.66   & 95.87 &93.6 &98.5 &82.9 &87.2 &61.8 &46.1 &41.7 &82.0 &37.5 &26.7 &45.9 &49.5 &60.0 &85.1 &67.3 &38.0 &89.2 &66.3\\
&Render PoseRNN              &32.61  & 73.1 & - & 91.7 &50.4 &62.1 &16.9 &6.6 &5.8 &30.5 &8.9 &6.7 &10.1 &16.3 &22.2 &70.6 &29.4 &20.2 &73.5 & - \\
&SegCNN w/o Pose             &68.35   & 95.61 &94.2 &98.6 &83.8 &89.5 &69.3 &47.5 &52.9 &83.9 &52.2 &43.5 &46.3 &52.9 &66.9 &87.0 &69.2 &40.0 &88.6 &63.8 \\
&SegCNN w pose GT            &79.37   & 97.1  &96.1 &99.4 &92.5 &93.9 &81.4 &68.8 &71.4 &90.8 &71.7 &64.2 &69.1 &72.2 &83.7 &91.3 &76.2 &58.9 &91.6 &56.7 \\
&SegCNN w Pose CNN           &68.6  & 95.67  &94.5 &98.7 &84.3 &89.3 &69.0 &46.8 &52.9 &84.9 &53.7 &39.5 &48.8 &50.4 &67.9 &87.5 &69.9 &42.8 &88.5 &60.9 \\
&SegCNN w Pose RNN &\textbf{69.93}  &\textbf{95.98} &
                                             94.9 &98.8 &85.3 &90.2 &71.9 &45.7 &57.0 &85.9 &58.5 &41.8 &51.0 &52.2 &69.4 &88.5 &70.9 &48.0 &89.3 &59.5 \\
\toprule[0.1 em]
\end{tabular}
\vspace{-0.8\baselineskip}
\begin{tabular}{llcccccccccccccccccccccc}
\vspace{-2.0\baselineskip}
Data& \multicolumn{1}{c}{Method} & \rotf{mIOU} & \rotf{Pix. Acc}  & \rotf{sky} & \rotf{car-lane} & \rotf{ped-lane}  & \rotf{$t$-stack} & \rotf{$t$-fence} & \rotf{wall} & \rotf{light-pole} & \rotf{$t$-light} & \rotf{tele-pole} & \rotf{$t$-sign} & \rotf{billboard} & \rotf{building} & \rotf{plants} & \rotf{car} & \rotf{cyclist} & \rotf{motorbike} & \rotf{truck} & \rotf{bus}\\
\hline
\parbox[t]{2mm}{\multirow{3}{*}{\rotatebox[origin=c]{90}{Dlake}}} & 
SegCNN w/o Pose  &62.36 &96.7 &95.3 &96.8 &12.8 &21.5 &81.9 &53.0 &44.7 &65.8 &52.1 &87.2 &55.5 &66.8 &94.5 &84.9 &20.3 &28.9 &78.4 &82.1 \\
&SegCNN w pose GT &73.10 &97.7 &96.8 &97.5 &41.3 &54.6 &87.5 &70.5 &63.4 &77.6 &70.5 &92.1 &69.2 &77.4 &96.1 &87.4 &24.5 &43.8 &80.0 &85.7 \\
&SegCNN w pose RNN &\textbf{67.00} &\textbf{97.1} &95.8 &97.2 &30.0 &37.4 &84.2 &62.6 &47.4 &65.5 &62.9 &89.6 &59.0 &70.3 &95.2 &86.8 &23.9 &34.4 &76.8 &86.6 \\
\toprule[0.1 em]
\end{tabular}
\caption{Compare the accuracy of different segment networks setting over Zpark (top) and Dlake (bottom) dataset. $t$ is short for 'traffic' in the table. Here we drop the 10 times S.D. to save space because it is relatively small. Our results are especially good at parsing of detailed structures and scene layouts, which is visualized in~\figref{fig:results}.}
\label{tbl:segment}
\vspace{-1.2\baselineskip}
\end{table*}

\textbf{Segment Evaluation.}
At top part of \tabref{tbl:segment}, we show the scene parsing results of Zpark dataset. 
Firstly, we adopt one of the SOTA parsing network on the CityScapes, \ie~ResNet38~\cite{WuSH16e}, and train it with Zpark dataset. It utilizes pre-trained parameters from the CityScapes~\cite{Cordts2016Cityscapes} dataset, and run with a 1.03$s$ per-frame with our resolution. 
As shown at the 1st row, it achieve reasonable accuracy compare to our segment CNN (2nd row) when there is no pose priori. However, our network is 10x faster. 
At 3rd row, we show the results of rendered label map with the estimated pose from pose RNN. Clearly, the results are much worse due to missing pixels and object misalignment.
At 4th row, we use the rendered label map with ground truth pose as segment CNN guidance to obtain an upper-bound for our segmentation performance. 
In this case, the rendered label map aligns perfectly with the image, thus significantly improves the results by correct labelling most of the static background.
At 5th and 6th row, we show the results trained with rendered label map with pose after pose CNN and pose RNN respectively. We can see using pose CNN, the results just improve slightly compare to the segment CNN. From our observation, this is because the offset is still significant for some detailed structures, \eg light-pole.

However, when using the pose after RNN, better alignment is achieved, and the segment accuracy is improved significantly especially for thin structured regions like pole, as visualized in~\figref{fig:results}, which demonstrates the effectiveness of our strategy. We list the results over Dlake dataset with more object labelling at bottom part of \tabref{tbl:segment}, and here the rendered label provides a background context for object segmentation, which also improve the object parsing performance. 


In \figref{fig:results}, we visualize several examples from our results at the view of camera. In the figure, we can see the noisy pose (a), is progressively rectified by pose CNN (b) and pose RNN (c) from view of camera. 
Additionally, at (d) and (e), we compare the segment results without and with camera pose respectively. As can be seen at the boxed regions, the segment results with rendered label maps provide better accuracy in terms of capturing region details at the boundary, discovering rare classes and keeping correct scene layout. All of above could be important for applications, \eg figuring out the traffic signs and tele-poles that are visually hard to detect.

\vspace{-0.8\baselineskip}
\section{Conclusion}
\vspace{-0.5\baselineskip}
\label{sec:conclusion}
In this paper, we present a deep learning based framework for camera self-localization and scene parsing with a given 3D semantic map for online videos, for the applications of visual-based outdoor robotic navigation. The algorithm fuses multi-sensors, is simple and runs efficient, meanwhile yields strong results in both of the tasks. More importantly, in our system, we show that the two information are mutually beneficial, where pose helps give good priori for segmentation and semantic guides a better localization. To perform the experiments, we created a dataset which contains a point-cloud based semantic 3D map and videos with ground truth camera pose and per-frame semantic labelling. 
{\small
\bibliographystyle{ieee}
\bibliography{egbib}
}

\end{document}